\documentclass[10pt,twocolumn,letterpaper]{article}

\usepackage[pagenumbers]{cvpr} 

\usepackage[dvipsnames]{xcolor}

\usepackage{amsmath,amsfonts,bm}

\newcommand{\T}{\mathcal{T}}    
\newcommand{\Ss}{\mathcal{S}}   
\newcommand{\R}{\mathbb{R}}     
\newcommand{\E}{\mathbb{E}}     

\newcommand{\xs}{\tilde{x}}     
\newcommand{\ys}{\tilde{y}}     
\newcommand{\ls}{\tilde{l}}     

\newcommand{\argmin}{\mathop{\arg\min}}

\newcommand{\loss}{\mathcal{L}}    
\newcommand{\ce}{\mathrm{CrossEntropy}}

\newcommand{\enc}{\mathcal{E}}  
\newcommand{\dec}{\mathcal{D}}  

\newcommand{\thetas}{\tilde{\theta}}

\definecolor{cvprblue}{rgb}{0.21,0.49,0.74}
\usepackage[pagebackref,breaklinks,colorlinks,citecolor=cvprblue]{hyperref}

\def\paperID{6266} 
\def\confName{CVPR}
\def\confYear{2024}

\title{Dataset Distillation in Latent Space}

\author{Yuxuan Duan, Jianfu Zhang, Liqing Zhang\thanks{Corresponding author.}\\
MoE Key Lab of Artificial Intelligence,
Shanghai Jiao Tong University, Shanghai, China \\
{\tt\small sjtudyx2016@sjtu.edu.cn, c.sis@sjtu.edu.cn, zhang-lq@cs.sjtu.edu.cn}
}

\usepackage{xspace}
\makeatletter
\DeclareRobustCommand\onedot{\futurelet\@let@token\@onedot}
\def\@onedot{\ifx\@let@token.\else.\null\fi\xspace}

\def\eg{\emph{e.g}\onedot} 
\def\ie{\emph{i.e}\onedot}

\def\wrt{w.r.t\onedot}

\makeatother

\usepackage[capitalize]{cleveref}
\Crefname{section}{Section}{Sections}
\Crefname{table}{Table}{Tables}
\Crefname{figure}{Figure}{Figures}
\Crefname{algorithm}{Algorithm}{Algorithms}

\usepackage{multirow}
\usepackage[ruled,linesnumbered,vlined]{algorithm2e}

\begin{document}
\maketitle

\begin{abstract}
Dataset distillation (DD) is a newly emerging research area aiming at alleviating the heavy computational load in training models on large datasets. It tries to distill a large dataset into a small and condensed one so that models trained on the distilled dataset can perform comparably with those trained on the full dataset when performing downstream tasks. Among the previous works in this area, there are three key problems that hinder the performance and availability of the existing DD methods: high time complexity, high space complexity, and low info-compactness. In this work, we simultaneously attempt to settle these three problems by moving the DD processes from conventionally used pixel space to latent space. Encoded by a pretrained generic autoencoder, latent codes in the latent space are naturally info-compact representations of the original images in much smaller sizes. After transferring three mainstream DD algorithms to latent space, we significantly reduce time and space consumption while achieving similar performance, allowing us to distill high-resolution datasets or target at greater data ratio that previous methods have failed. Besides, within the same storage budget, we can also quantitatively deliver more latent codes than pixel-level images, which further boosts the performance of our methods. 
\end{abstract}
\section{Introduction}
\label{sec:intro}

\begin{figure*}[t]
    \centering
    \includegraphics[width=.7\linewidth]{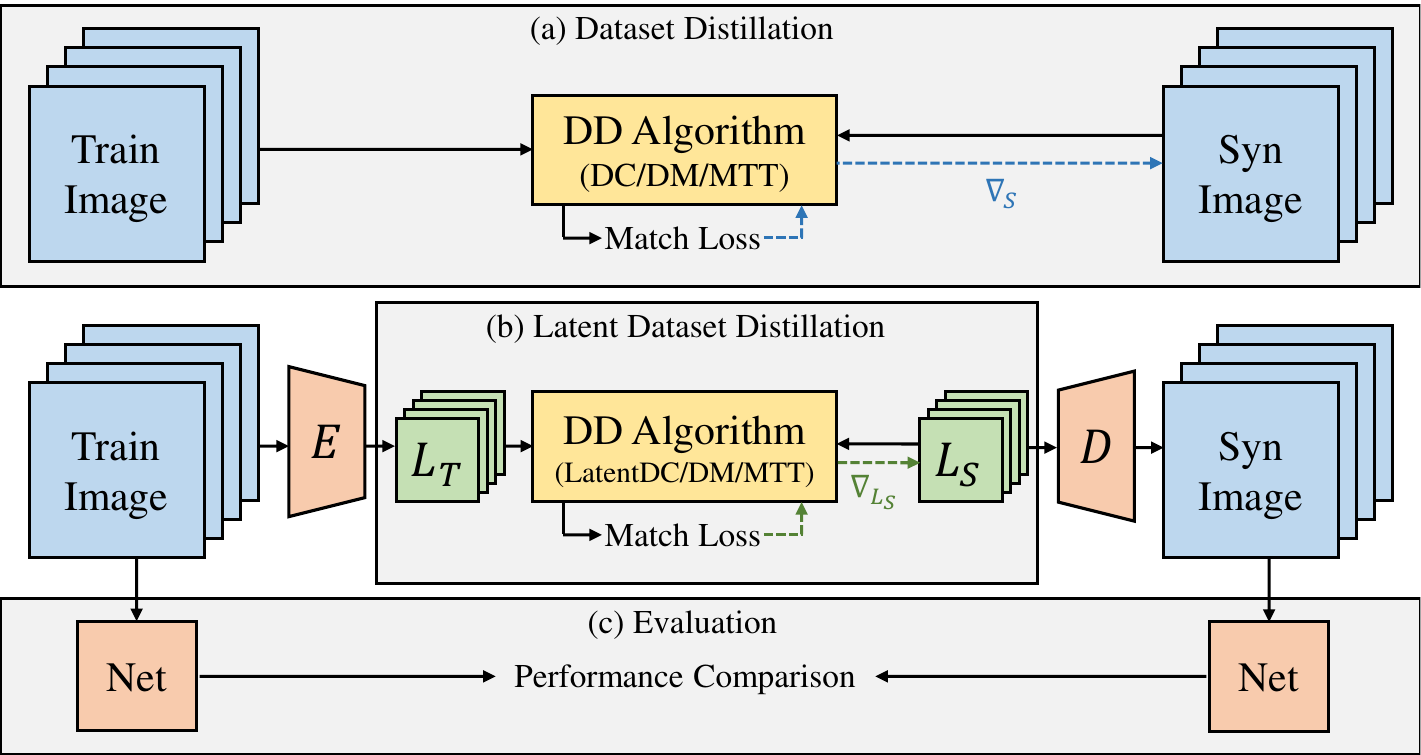}
    \caption{An overview of dataset distillation (DD) and LatentDD. (a) Procedure of DD in pixel space, where DD algorithms produce gradients to update synthetic images $\Ss$. (b) Procedure of DD in latent space, where DD algorithms directly operate on latent codes and produce gradients to update synthetic latent codes in $L_\Ss$, without decoding latent codes into images before feeding them into DD algorithms. (c) After distillation, networks trained on real training images $\T$ and synthetic images $\Ss$ will be compared.}
    \label{fig:intro}
\end{figure*}

Due to the rapidly progressing computation capability of modern devices, people are building unprecedentedly large and data-hungry models. For instance, the state-of-the-art text-to-image generative models, Stable Diffusions \citep{ldm}, were pretrained on LAION-5B \citep{laion5b}, which contains 5.85 billion text-image pairs. Although large models trained on large datasets have achieved fascinating performance in many applications, they are still known for the high demands on training time, computing devices, storage budgets, and electricity consumption.

In recent years, dataset distillation (DD) has become a newly emerging research topic which tries to lower the aforementioned demands. Inspired by knowledge distillation \citep{kd1,kd2,kd3,kd4}, DD aims to distill a full dataset into a much smaller synthetic set, so that in specific tasks the models trained on this distilled dataset are expected to perform comparably to those trained on the full dataset. \citet{dd} proposed a pioneering prototype method of DD by solving a bi-level optimization problem, which has inspired many following works. By analyzing these works, we summarize the common issues in DD into three problems. 
\textbf{Problem 1:} Most DD methods involve solving a computationally intensive optimization problem, which has high time complexity. 
\textbf{Problem 2:} Most methods have to store the computation graph recording the optimization process of the network before updating the distilled dataset
, making its space complexity high too, especially for those bi-level optimization methods with nested loops.
\textbf{Problem 3:} As we expect to remain a small data ratio between the size of distilled dataset and full dataset, the distilled dataset should be highly info-compact to cover as much information as possible. However, distilling dataset in the original space (\eg pixel space for images) will inevitably condense high-frequency details into limited storage budget, which are usually less necessary for downstream tasks.

Methods following the prototype of DD \citep{dd} have usually focused on one or two of the three problems. For example, the three mainstream DD algorithms DC \citep{dc}, DM \citep{dm} and MTT \citep{mtt} respectively use gradient matching, feature matching and parameter matching to efficiently approximate the bi-level optimization of the prototype, improving P1--P2 yet omitting P3. Some other works factorize the distilled dataset into more info-compact \emph{components} \citep{idc,haba,glad} thus alleviate P3. Nevertheless, since these methods still operate on pixel space using the DD algorithms above, they even induce extra time and space consumption when transforming the components to images and inversely back-propagating gradients from images to components, thus fail in P1--P2. To the best of our knowledge, no previous work simultaneously settles all the three problems.

Within the field of image generation, another domain fraught with computational complexity, recent breakthroughs in diffusion models \citep{ddpm,ddim,adm} have pushed both the performance and the time \& space consumption to a new level. Later on, \citet{ldm} have proposed latent diffusion models transferring the diffusing/denoising procedures from pixel space to latent space with the help of a pretrained autoencoder, which largely accelerates the training process while keeping the performance. Inspired by such design, we propose \textbf{Latent Dataset Distillation (LatentDD)} which transfers the three mainstream DD algorithms DC, DM and MTT to fully operate on latent codes encoded by the pretrained generic autoencoder provided in latent diffusion models rather than on images, namely LatentDC, LatentDM and LatentMTT. An overview of LatentDD is shown in \cref{fig:intro} (b) and (c). Since the latent codes have much smaller size than pixel-level images, LatentDD takes significantly less time and space (both main memory and GPU memory) to run DD algorithms, alleviating P1--P2. Such acceleration and space reduction is only at the cost of marginal performance degradation, as the pretrained autoencoder can roughly keep the distribution of the original images into the latent codes and thus solving DD in latent space is approximately equivalent to that in pixel space (see \cref{sec:latentspace}). As for P3, the latent codes are naturally info-compact representations of the original images since the autoencoder can reconstruct the images from latent codes losing just the subtlest details. Besides, with a fixed data ratio (storage budget) in DD tasks, we can quantitatively store much more latent codes than pixel-level images, which boosts the performance of LatentDD.

In summary, our work is the first to settle all the three problems in DD at the same time. Additionally, it is noteworthy that its fast and space-saving designs enable LatentDD to distill high-resolution datasets. While most of the recent works have been dealing with toy datasets like CIFAR10/100 \citep{cifar} while only a few latest ones trying higher resolution like 64 or 128, we roll out our experiments on high-resolution settings starting from 256, and beyond. These challenging experiments in \cref{sec:experiment} have manifested the superiority of LatentDD over previous works.

\section{Related Work}
\label{sec:related}

When selecting representatives from large-scale datasets, coreset selection \citep{coreset1,coreset2,coreset3,coreset4,coreset5,coreset6,coreset7,dq} was once the primary solution. Since \citet{dd}, follow-up works started to focus on dataset distillation, aiming at solving the three problems mentioned in \cref{sec:intro}. 
Besides the brief introductions below, some surveys \citep{survey1,survey2,survey3,survey4} are also recommended for more details.

\paragraph{Gradient Matching} 

\citet{dc} proposed Dataset Condensation (DC), the first practically plausible DD algorithm which simplified the clumsy bi-level optimization of the prototype. DC used single-step gradient matching as a surrogate objective to bridge the parameter gap when trained on real/synthetic datasets. Following DC, DSA \citep{dsa} attached Differentiable Siamese Augmentation to DC framework. DCC/DSAC \citep{dcc} enhanced DC/DSA with contrastive signal, matching gradient in an all-class manner instead of a class-wise one.

\paragraph{Feature Matching}

\citet{dm} proposed Distribution Matching (DM), which extracted features from both real/synthetic images via randomly initialized networks and matched their mean values class-wisely. As another mainstream DD algorithm, DM largely accelerated DD process by completely avoiding bi-level optimization. IDM \citep{idm} improved DM with image partitioning \cite{idc} and trained feature extractors. Similar to DM, CAFE \citep{cafe} also compared the mean values of multi-layer features, yet equipped with an extra discrimination loss. 

\paragraph{Parameter Matching}

\citet{mtt} proposed the third DD algorithm Matching Training Trajectories (MTT). It reduced the accumulated parameter error in gradient matching methods by matching model parameters after relatively long-term training trajectories. After MTT, \citet{paramprune} pruned hard-to-match parameters, FTD \citep{ftd} regularized flat trajectories during the buffer phase, and TESLA \citep{tesla} improved MTT by lowering its space complexity. 

\paragraph{Optimization}

Besides designing DD algorithms, some other works attempted to enhance these algorithms with optimization techniques, including kernel method \citep{krr,kip,rfad,frepo}, label learning \citep{ld,tesla}, model augmentation \citep{ma}, clustering \citep{dream} and calibration \citep{calibration}.

\paragraph{Factorization}

Factorization is another research direction orthogonal to designing DD algorithms. It aims at factorizing distilled images into more info-compact components, so that these components can be transformed into quantitatively more images for downstream tasks than directly storing pixel-level images within the same storage budget. Specific strategies include image partitioning \citep{idc,idm} and factorizing images into latent codes + decoders \citep{haba,rtp,kfs,glad}. However, all these methods had to repeatedly restore the pixel-level images before sending them into DD algorithms, and back-propagate the gradients from the images to the components, resulting in heavy time \& space overhead. On the contrary, as a factorization method, our LatentDD instead directly operates in latent space, which largely reduces time \& space consumption.

\section{Latent Dataset Distillation}
\label{sec:latentdd}


\subsection{Problem Definition}
\label{sec:definition}

Suppose we have a large dataset $\T = \{(x_i, y_i)\}_{i=1}^{|\T|}$ to be distilled, which consists of real pairs of datum $x_i \in \R^d$ and class label $y_i \in \{0, \dots, C - 1\}$ where $d$ is the dimension of the data and $C$ is the number of classes. The goal is to seek a distilled dataset $\Ss = \{(\xs_i, \ys_i)\}_{i=1}^{|\Ss|}$ including synthetic pairs of datum $\xs_i$ and label $\ys_i$, and $|\Ss| \ll |\T|$. Conventionally the synthetic data follow the form of real data (\ie $\xs_i \in \R^d$ and $\ys_i \in \{0, \dots, C - 1\}$). However, there are also some works exploring more effective forms of data via factorization, or labels via label learning, as long as $\Ss$ does not exceed a predefined storage budget.

With the distilled $\Ss$, we expect that models trained on it will achieve comparable performance with those trained on the real dataset $\T$ in downstream tasks. Formally,
\begin{equation}
    \small
    \begin{aligned}
    \Ss^* = \argmin_{\Ss} \|\loss^\T(\theta^\Ss) - \loss^\T(\theta^\T)\| \quad \text{subject to} \\
    \theta^\Ss = \argmin_\theta \loss^\Ss(\theta), \quad \theta^\T = \argmin_\theta \loss^\T(\theta),
    \end{aligned}
\end{equation}
where $\theta^\Ss$ and $\theta^\T$ are models trained on $\Ss$ and $\T$, and $\loss^\Ss$, $\loss^\T$ are respectively a certain objective (loss function) when evaluated on the two datasets. Since models trained on $\Ss$ are unlikely to outperform those trained on $\T$, we usually seek $\Ss^*$ that performs the best:
\begin{equation}
    \small
    \Ss^* = \argmin_{\Ss} \loss^\T(\theta^\Ss) \quad \text{subject to} \quad \theta^\Ss = \argmin_\theta \loss^\Ss(\theta).
\label{eq:dd}
\end{equation}

\subsection{From Pixel to Latent Space}
\label{sec:latentspace}

The prototype of DD straightforwardly solves \cref{eq:dd} in a bi-level optimization manner, and later three mainstream DD algorithms DC, DM and MTT have been proposed to efficiently solve \cref{eq:dd} by optimizing surrogate objectives. Primarily focusing on image classification, all the previous works have distilled datasets in pixel space. When solving \cref{eq:dd} with DD algorithms, data in both $\T$ and $\Ss$ are in the form of the original pixel-level images (\ie $x_i, \xs_i \in \R^{3 \times H \times W}$, as three-channel images with size $(H, W)$). Although some previous works have proposed factorization methods which distill more info-compact components instead of images, they are still based on DD algorithms in pixel space. As pixel-level images usually consist of low-frequency information, which includes the contents that are truly necessary to downstream tasks like image classification, \emph{and} high-frequency information including fine and subtle details and even noises. Taking both parts of information into account when running DD algorithms has brought considerable overhead in time and space consumption, especially when distilling high-resolution datasets or aiming at a higher data ratio. Also, with the useless high-frequency information occupying some of the storage budget, the less info-compact synthetic images fail to reach better scores in classification tasks.

In this work, we attempt to transfer the DD algorithms to directly operate in latent space rather than pixel space. Suppose we have an autoencoder $\{\enc(\cdot), \dec(\cdot)\}$ pretrained on a large dataset to ensure its generalization ability to encode and reconstruct any image with its encoder $\enc(\cdot)$ and decoder $\dec(\cdot)$. Specifically, from a real image $x_i \in \R^{3 \times H \times W}$ in $\T$, the encoder can encode it into a latent code $l_i \in \R^{C \times \frac{H}{f} \times \frac{W}{f}}$ as a $C$-channel feature map with a downsampling factor $f$, and the settings of $C$ and $f$ ensure that the size of $l_i$ is smaller than $x_i$ (\ie $C < 3\cdot f^2$). Then, from a latent code $l_i$, the decoder can decode it into a reconstructed image $x_i' \in \R^{3 \times H \times W}$ back to the original size. If the autoencoder is well trained, the reconstructed image $x_i'$ should be roughly the same as the original image $x_i$, with acceptable minor loss of details and noises. In this way, the latent code $l_i$ can naturally serve as an info-compact representation of the original image $x_i$ since most of the necessary information needed to reconstruct $x_i$ has been encoded into $l_i$. Therefore, if we can use latent codes instead of images in DD algorithms, it will improve DD processes by simultaneously alleviate the three problems mentioned in \cref{sec:intro}.

\begin{figure}[t]
    \centering
    \begin{minipage}[c]{0.49\linewidth}
        \includegraphics[width=\linewidth]{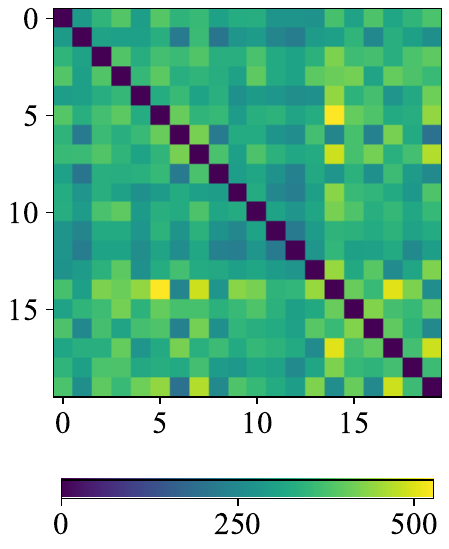}
    \end{minipage}\hspace{1mm}%
    \begin{minipage}[c]{0.49\linewidth}
        \includegraphics[width=\linewidth]{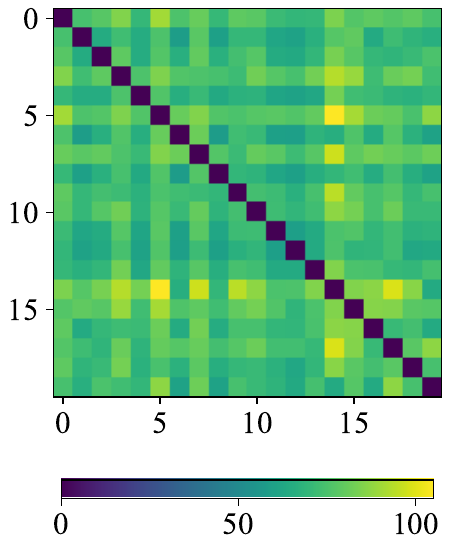}
    \end{minipage}
    \caption{The Euclidean distance matrices of 20 randomly sampled images (left) from subset \emph{Bird} of ImageNet and their corresponding latent codes (right).}
        \label{fig:distmat}
\end{figure}

Stable Diffusion, the most widely used instance of latent diffusion models \cite{ldm}, is equipped with such a pretrained generic autoencoder, which successfully transfer the time-consuming denoising processes to latent space. In our LatentDD, we utilize this off-the-shelf autoencoder as a converter between images in pixel space and latent codes in latent space. By default, this autoencoder is capable of encoding any image with any resolution into a latent code with $C = 4$ channels and a downsampling factor $f = 8$. For instance, an RGB image of resolution $512 \times 512$ will be encoded into a latent code of size $4 \times 64 \times 64$. By these settings, a latent code is only $1/48$ of the original image \wrt the number of parameters, which is highly info-compact. 

Nevertheless, before transferring DD algorithms to latent space, we still need to verify that this autoencoder will not break the original distribution of the images, as keeping this distribution is critical to image classification tasks. In a preliminary experiment, we randomly select 20 images from \textit{Bird}, a subset of ImageNet (see \cref{sec:setting} for details), encode them into latent codes, and respectively show the Euclidean distance matrices of these images and latent codes in \cref{fig:distmat}. From the two heatmaps, we may conclude that the latent codes approximately remain the distribution of the original images, as the relative distances among these latent codes strongly correlate with those among the images. As a result, the learned classification hyperplanes in latent space can correspond to the hyperplanes in pixel space, making DD in latent space and pixel space approximately equivalent. Such equivalence is also empirically validated by ablation study in \cref{sec:ablation}, where training classifiers on the full sets of images/latent codes (\cref{tab:full}) and distilling the datasets into same number of images/latent codes (\cref{tab:ipc1vslpc1}) both render similar performance. 

Based on \cref{eq:dd}, DD in latent space can be written as
\begin{equation}
    \small
    L_\Ss^* = \argmin_{L_\Ss} \loss^{L_\T}(\theta^{L_\Ss}) \quad \text{subject to} \quad \theta^{L_\Ss} = \argmin_\theta \loss^{L_\Ss}(\theta),
\label{eq:latentdd}
\end{equation}
where $L_\T = \{(\enc(x_i), y_i)\}_{i=1}^{|\T|}$ and $L_\Ss = \{(\ls_i, \ys_i)\}_{i=1}^{|\Ss|}$ are respectively the set of real and synthetic pairs of latent code and its label, and we remain $\ys_i \in \{0, \dots, C - 1\}$ as class labels. 
Before running LatentDD algorithms, $L_\T$ is precomputed and stored in the main memory for fast retrieval, and $L_\Ss$ is initialized with randomly selected real latent codes. After obtaining $L_\Ss^*$ via LatentDD, we can reconstruct the synthetic image dataset as $\Ss^* = \{(\dec(\ls_i), \ys_i)\}_{i=1}^{|\Ss|}$ for downstream tasks, as depicted in \cref{fig:intro}.

As previous works \citep{haba,idc,rtp} focusing on factorization have proved, the fixed data ratio will severely confine the performance on downstream image classification tasks if we stick to delivering distilled datasets as pixel-level images within a limited storage budget. These works instead deliver info-compact components such as low-resolution thumbnails or a combination of latent codes and decoders within the storage budget, which can be resized or decoded into more images than the same budget can directly store. Our LatentDD has followed this idea, as we deliver $n\cdot 3f^2/C$ latent codes rather than $n$ pixel-level images. It is also worth mentioning that, previous works of latent codes + decoders also train their decoders along with the latent codes during DD processes, thus their decoders can only be exclusively used on specific datasets (or even specific classes of these datasets) and specific resolutions. So they have to store the decoders as a part of the distilled datasets to be delivered. On the contrary, since our pretrained autoencoder is generic for any image and any resolution, and is also publicly available online, it only takes $O(1)$ space to store the decoder when we distill $N$ datasets instead of $O(N)$ as in previous works. Hence our decoder will averagely take negligible space if we distill many datasets and may be excluded from the storage budget like an unparameterized resizing operation that can be applied to any image.

\subsection{Latent Dataset Distillation Algorithms}
\label{sec:algo}

Dataset Condensation (DC), Distribution Matching (DM) and Matching Training Trajectories (MTT) are three mainstream DD algorithms, which solve the DD problem in \cref{eq:dd} with surrogate objectives of gradient matching, feature matching and parameter matching respectively. We show how to seamlessly transfer these algorithms to latent space as below. Pseudocodes providing detailed procedures are available in Appendix.

\subsubsection{LatentDC}
\label{sec:latentdc}

DC \citep{dc} is designed based on the observation that, if the model $\theta^\Ss$ trained on distilled dataset $\Ss$ has similar parameters with $\theta^\T$ trained on real dataset $\T$, it will be certain to perform comparably when evaluated on test set. Such resembling parameters can be achieved by matching the gradients of network parameters $\nabla_\theta$ induced by training the same model $\theta_t$ (at timestep $t$) on $\Ss$ and $\T$. Formally,
\begin{equation}
    \small
    \begin{aligned}
    \Ss^* = \argmin_{\Ss} \E_{\theta_0 \sim P_{\theta_0}}[\sum_{t=0}^{T-1} D(\nabla_\theta \loss^\Ss(\theta_t), \nabla_\theta \loss^\T(\theta_t))] 
    \\ 
    \text{subject to} \quad \theta_{t+1} \gets \theta_{t} - \eta \nabla_\theta \loss^\Ss(\theta_{t}),
    \end{aligned}
\label{eq:dc}
\end{equation}
where DC randomly initializes network parameters $\theta_0$ and repeatedly matches the two groups of gradients with a cosine-like gradient matching loss $D(\cdot, \cdot)$ along the $T$-step training process of $\theta_t$ on synthetic dataset $\Ss$ and classification criterion $\loss$ (usually a cross-entropy loss).

\begin{figure}[t]
    \centering
    \includegraphics[width=.9\linewidth]{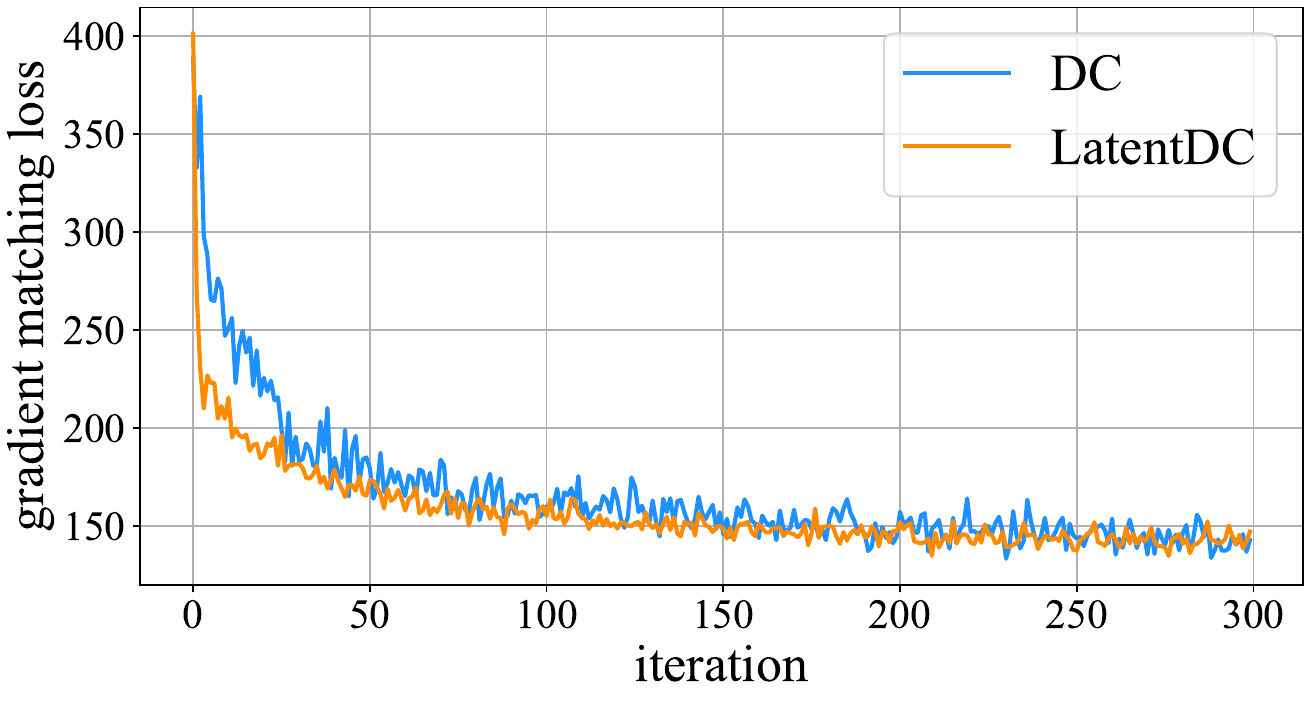}
    \caption{MSE gradient matching loss of DC and LatentDC in the first 300 iterations when distilling subset \textit{Bird} of ImageNet.}
    \label{fig:lossplot}
\end{figure}

To run DC in latent space instead of pixel space, the most essential modification is that we have to replace the network $\theta$ operating on pixel-level images with a new network $\thetas$ operating on latent codes. Since the latent codes $l \in \R^{C \times \frac{H}{f} \times \frac{W}{f}}$ still remain a 2D spatial structure just as images, commonly used convolutional architectures are also applicable to them if only we accordingly change the channel numbers and feature map sizes. Besides, as the size of $l$ is much smaller than images, we can use lighter networks with fewer layers as $\thetas$. For instance, while distilling image datasets of resolution 256 conventionally uses ConvNetD6 with depth 6, we apply ConvNetD3 as $\thetas$ when $f = 8$ and ConvNetD4 when $f = 4$, just as dealing with images of the corresponding 
resolution (refer to \cref{sec:setting} for details). Being able to reduce the size of networks used in DD is one of the main reasons that our latent version of DD algorithms takes much less time and space to run. With $\thetas$, our LatentDC can be formulated as
\begin{equation}
    \small
    \begin{aligned}
    L_\Ss^* = \argmin_{L_\Ss} \E_{\thetas_0 \sim P_{\thetas_0}}[\sum_{t=0}^{T-1} D(\nabla_{\thetas} \loss^{L_\Ss}(\thetas_t), \nabla_{\thetas} \loss^{L_\T}(\thetas_t))] \\
    \text{subject to} \quad \thetas_{t+1} \gets \thetas_{t} - \eta \nabla_{\thetas} \loss^{L_\T}(\thetas_{t}),
    \end{aligned}
\label{eq:latentdc}
\end{equation}
where we follow \citet{idc} to (1) update $\thetas$ using real latent codes in $L_\T$ instead of synthetic ones in $L_\Ss$ as the gradients will quickly vanish if trained on the latter and (2) use an MSE-like gradient matching loss for $D(\cdot, \cdot)$ which better fits training on real datasets. In \cref{fig:lossplot}, we illustrate the gradient matching losses of both DC and LatentDC in the first 300 iterations when distilling \emph{Bird} into one image/latent code per class (see \cref{sec:ablation}), where the loss of LatentDC decreases slightly faster, and more steadily than its counterpart in pixel space in the beginning. After that, DC and LatentDD both converge to a similar level of loss, which matches the resembling performance on downstream classfication tasks reported in \cref{tab:ipc1vslpc1}.

Based on DC, \citet{dsa} add Differentiable Siamese Augmentation (DSA) during both training and evaluation stages, which has become a standard technique in following works. However, according to our preliminary experiments, only two transformations (crop, cutout) used in DSA can also be applied to latent codes, while the others (color, scale, rotate, flip) on latent codes will unexpectedly affect the quality of the decoded images. Hence, during the training stage in \cref{eq:latentdc} we do not apply DSA, but still augmenting the pixel-level images decoded from latent codes in the evaluation stage (see Appendix). Such strategy is also adopted by LatentDM and LatentMTT. We suppose that designing a set of feasible transformations for augmenting latent codes is another topic worth researching on, yet we leave this part for future work.

\subsubsection{LatentDM}

Unlike the other two algorithms DC and MTT, DM \citep{dm} is designed aiming at totally eliminate the computationally intensive bi-level optimization. It updates the distilled dataset $\Ss$ so that the empirical estimate of maximum mean discrepancy (MMD) is minimized between each class of $\Ss$ and $\T$:
\begin{equation}
    \small
    \begin{aligned}
    & \Ss^* = \argmin_{\Ss} \E_{\theta \sim P_{\theta}} \\
    & [\sum_{c = 0}^{C - 1}    \|\frac{1}{|\T_c|} \sum_{(x_i,y_i) \in \T_c}\phi_\theta(x_i) - \frac{1}{|\Ss_c|}\sum_{(\xs_i,\ys_i) \in \Ss_c}\phi_\theta(\xs_i)\|^2],
    \end{aligned}
\label{eq:dm}
\end{equation}
where $\T_c$, $\Ss_c$ are the subsets of class $c$, and $\phi_\theta$ is an embedding function based on randomly initialized networks $\theta$. Similar to LatentDC, we can also transfer DM to LatentDM by using embedding networks $\thetas$ of proper architecture:
\begin{equation}
    \small
    \begin{aligned}
    & L_\Ss^* = \argmin_{L_\Ss} \E_{\thetas \sim P_{\thetas}} \\
    & [\sum_{c = 0}^{C - 1} \|\frac{1}{|L_{\T_c}|} \sum_{(l_i,y_i) \in L_{\T_c}}\phi_{\thetas}(l_i) - \frac{1}{|L_{\Ss_c}|}\sum_{(\ls_i,\ys_i) \in L_{\Ss_c}}\phi_{\thetas}(\ls_i)\|^2],
    \end{aligned}
\label{eq:latentdm}
\end{equation}
where the embedding function $\phi_{\thetas}$ now extracts embeddings from latent codes rather than images.

\subsubsection{LatentMTT}

MTT \citep{mtt} is designed to alleviate the issue of the accumulated parameter error caused by the difference between gradients $\nabla_\theta \loss^\Ss(\theta_t)$ and $\nabla_\theta \loss^\T(\theta_t)$ in \cref{eq:dc} of DC. Starting from the same network $\theta_t^\T$ that has been trained on the real dataset $\T$ for $t$ steps, it matches the parameters of two networks $\theta_{t + N}^{\Ss}$ and $\theta_{t + M}^{\T}$ respectively after a student trajectory which trains on $\Ss$ for $N$ steps and an expert trajectory which trains on $\T$ for $M$ steps:
\begin{equation}
    \small
    \Ss^* = \argmin_{\Ss} \E_{t \in \{0, \dots, T^+\}} \left[\frac{\|\theta_{t + N}^{\Ss} - \theta_{t + M}^{\T}\|_2^2}{\|\theta_t^{\T} - \theta_{t + M}^{\T}\|_2^2}\right],
    \label{eq:mtt}
\end{equation}
where the starting step $t$ is sampled within the limit of a maximum starting step $T^+$ since the later part of the a training trajectory is less informative. By matching parameters through multi-step trajectories instead of single-step gradients, MTT generally outperforms DC at the expense of greater time and space consumption. Just as in LatentDC and LatentDM, we can modify MTT into LatentMTT by moving both the expert and the student trajectories into the latent space:
\begin{equation}
    \small
    L_\Ss^* = \argmin_{L_\Ss} \E_{t \in \{0, \dots, T^+\}} \left[\frac{\|\thetas_{t + N}^{L_\Ss} - \thetas_{t + M}^{L_\T}\|_2^2}{\|\thetas_t^{L_\T} - \thetas_{t + M}^{L_\T}\|_2^2}\right],
    \label{eq:latentmtt}
\end{equation}
where we also pre-buffer the expert trajectories in latent space for faster run-time matching, similar to MTT in pixel space.
\section{Experiments}
\label{sec:experiment}

\subsection{Experimental Settings}
\label{sec:setting}
\paragraph{Datasets}

To fully illustrate the capability of LatentDD, we conduct experiments on high-resolution datasets starting from 256, which is higher than the maximum resolution of almost all the previous works. We omit the less challenging low-resolution datasets such as MNIST and CIFAR since these datasets can be managed by most previous methods. Specifically, we take five subsets of ImageNet \citep{imagenet}, namely \emph{Bird} (ImageSquawk), \emph{Fruit} (ImageFruit), \emph{Woof} (ImageWoof), \emph{Cat} (ImageMeow) and \emph{Nette} (ImageNette), where Woof and Nette are online resources\footnote{\url{https://github.com/fastai/imagenette}} and the other three comes from \citet{mtt}. Our experiments cover different settings of DD aiming at image classification as the downstream task, including resolution 256 or 512, and image per class (IPC) 1 or 10. See Appendix for the categories included in these ImageNet subsets and the procedures of preprocessing the images.

\paragraph{Baselines}

We select previous state-of-the-art methods based on each of the three mainstream DD algorithms DC, DM and MTT into our baseline list. For DC, we include DSA \citep{dsa} with augmentation, IDC \citep{idc} which partitions images and GLaD DC \citep{glad} based on GAN prior. For DM, we include the original DM \citep{dm} and GLaD DM \cite{glad}. Finally for MTT, we include MTT \citep{mtt}, GLaD MTT \citep{glad} and FTD \citep{ftd} with flattened expert trajectories. Since no previous work has reported performance under these challenging settings, the results are based on our evaluation using their public implementations with as little modification as possible to adapt to high-resolution scenarios.

\paragraph{Choice of Autoencoder}

Although we may use any autoencoder that remains the distribution of the original images among the latent codes, we practically choose the one as a part of Stable Diffusion v1.4\footnote{\url{https://huggingface.co/CompVis/stable-diffusion-v-1-4-original}} as the pretrained generic autoencoder. This autoencoder is capable of encoding any image with a downsampling factor $f = 8$. Though this autoencoder can encode images of any resolution, it was originally trained on resolution 512 and we have observed that more details will be lost in the reconstructed images if used on lower resolution. Therefore, we adopt a preprocessing procedure, in which we first upsample the original pixel-level dataset $\T$ by a factor of two, and then encode the resized dataset into latent codes $L_\T$, resulting in a downsampling factor $f = 4$. Before evaluation, we also follow a postprocessing procedure where we downsample the decoded images by a factor of two, restoring the synthetic dataset $\Ss$ at the original resolution. By default, we apply these procedures of $f = 4$ in the experiments below unless otherwise specified. See Appendix for further information and analysis of autoencoder.

\paragraph{Implementation Details}
Please refer to Appendix for hyperparameter settings, training/evaluation protocols and pseudocodes for the three LatentDD algorithms.

\subsection{LatentDD vs. Baselines}
\label{sec:result}

\begin{table*}[t]
    \centering
    \setlength{\tabcolsep}{1.9mm}
    \begin{tabular}{ll|ccccc|cc|cc}
        \hline
        \multirow{3}{*}{Algo.} & \multirow{3}{*}{Method} & \multicolumn{7}{c|}{Res. 256} & \multicolumn{2}{c}{Res. 512} \\
        & & \multicolumn{5}{c|}{IPC 1} & \multicolumn{2}{c|}{IPC 10} & \multicolumn{2}{c}{IPC 1} \\
        & & Bird & Fruit & Woof & Cat & Nette & Bird & Fruit & Bird & Fruit \\
        \hline
        \multirow{4}{*}{DC} & DSA & 30.52 & 20.28 & 22.12 & 22.20 & 34.40 & 45.52 & 30.48 & 25.44 & 16.40 \\
        & IDC & 36.28 & 24.60 & 25.64 & 27.68 & 48.16 & 64.28 & 39.68 & 28.68 & 20.12 \\
        & GLaD DC & 30.32 & 19.56 & 21.84 & 22.24 & 34.44 & 46.04 & 32.60 & 25.80 & 16.72 \\
        & \textbf{LatentDC} & \textbf{46.72} & \textbf{30.12} & \textbf{28.96} & \textbf{38.08} & \textbf{55.92} & \textbf{80.44} & \textbf{51.60} & \textbf{47.52} & \textbf{29.68} \\
        \hline
        \multirow{3}{*}{DM} & DM & 27.64 & 19.48 & 20.04 & 21.16 & 32.08 & 41.56 & 28.12 & 28.28 & 19.72 \\
        & GLaD DM & 28.84 & 21.28 & 21.28 & 20.52 & 32.40 & - & - & 29.32 & 20.68 \\
        & \textbf{LatentDM} & \textbf{47.08} & \textbf{30.68} & \textbf{28.00} & \textbf{36.28} & \textbf{56.08} & \textbf{77.20} & \textbf{47.76} & \textbf{46.20} & \textbf{30.60} \\
        \hline
        \multirow{4}{*}{MTT} & MTT & 35.80 & 21.08 & 24.92 & 24.16 & 40.52 & - & - & - & - \\
        & GLaD MTT & 32.20 & 20.48 & 23.00 & 21.44 & 31.84 & - & - & - & - \\
        & FTD & 35.96 & 21.76 & 26.32 & 26.60 & 40.96 & - & - & - & - \\
        & \textbf{LatentMTT} & \textbf{52.86} & \textbf{37.82} & \textbf{39.84} & \textbf{41.42} & \textbf{62.86} & \textbf{78.44} & \textbf{52.46} & \textbf{52.44} & \textbf{36.20} \\
        \hline
    \end{tabular}
    \caption{Quantitative results of dataset distillation experiments on ImageNet subsets. LatentDD algorithms follow the setting of $f = 4$, where $\text{LPC} = 12 \times \text{IPC}$. The mean classification accuracy among five times of evaluations is reported. The results marked as a hyphen - indicate that the methods have run out of 24GB GPU memory during the experiments.}
    \label{tab:quan}
\end{table*}

\begin{table}[t]
\centering
\begin{tabular}{l|rr|r}
    \hline
    \multirow{2}{*}{Method} & \multicolumn{2}{c|}{Bird 256} & \multicolumn{1}{c}{Bird 512} \\
    & \multicolumn{1}{c}{IPC 1} & \multicolumn{1}{c|}{IPC 10} & \multicolumn{1}{c}{IPC 1} \\
    \hline
    Build & \multicolumn{2}{r|}{\textbf{2m} / 26.1GB} & \multicolumn{1}{r}{\textbf{2m} / 94.6GB} \\
    \textbf{Build Latent} & \multicolumn{2}{r|}{10m / \textbf{5.0GB}} & \multicolumn{1}{r}{52m / \textbf{9.9GB}} \\
    \hline
    DSA & 6h 34m & 30h 27m & 25h 17m \\
    IDC & 32h 36m & 40h 26m & 46h 50m \\
    GLaD DC & 15h 15m & 119h 20m & 30h 51m \\
    \textbf{LatentDC} & \textbf{1h 59m} & \textbf{3h 18m} & \textbf{3h 21m} \\
    \hline
    DM & 10m & 14m & 51m \\
    GLaD DM & 55m & - & 2h 15m \\
    \textbf{LatentDM} & \textbf{1m} & \textbf{2m} & \textbf{1m} \\
    \hline
    MTT Buffer & \multicolumn{2}{r|}{16h 20m} & 48h 50m \\
    FTD Buffer & \multicolumn{2}{r|}{24h 00m} & 92h 30m\\
    \textbf{Latent Buffer} & \multicolumn{2}{r|}{\textbf{50m}} & \textbf{3h 10m}\\
    \hline
    MTT & 7h 03m & - & - \\
    GLaD MTT & 11h 48m & - & -\\
    FTD & 7h 17m & - & - \\
    \textbf{LatentMTT} & \textbf{2h 14m} & \textbf{4h 16m} & \textbf{2h 38m} \\
    \hline
\end{tabular}
\caption{Time and space (main memory) consumption during the dataset building, expert trajectory buffering and training processes of LatentDD and the baselines, evaluated on a single NVIDIA RTX 4090. LatentDD algorithms follow the setting of $f = 4$, where $\text{LPC} = 12 \times \text{IPC}$. A hyphen - indicates that the method has run out of 24GB GPU memory during the experiment.}
\label{tab:timespace}
\end{table}

The quantitative results of the comparisons between LatentDD algorithms and baseline methods are shown in \cref{tab:quan}, where the distilled datasets are evaluated on ConvNets \cite{convnet} whose depths correspond to image resolutions (see Appendix). In these experiments, our LatentDD algorithms follow the previous factorization-based methods which deliver info-compact components instead of images \cite{idc,idm,haba,rtp,kfs}, delivering 12 latents per class (LPC) within the same storage of 1 IPC, and accordingly 120 LPC within 10 IPC under the setting of $f = 4$. As our LatentDD methods deliver highly info-compact latent codes, they are able to significantly outperform previous works which deliver less info-compact low-resolution thumbnails (IDC) or full-size images (other baselines). In this way, our LatentDD has successfully settled the info-compactness problem (P3) mentioned in \cref{sec:intro}. Additionally, we will show cross-architecture results and depict some distilled samples with three LatentDD algorithms in Appendix.

Besides, we also list the running time and main memory consumption of the above experiments in \cref{tab:timespace}. Though LatentDD methods spend more time on building the datasets into main memory since they pre-encode the real datasets into latent codes, the buffering processes of LatentMTT and the training processes of all latent methods have been largely accelerated due to the smaller sizes of both the latent codes and the networks. The time spent on decoding latent codes into images is negligible since it only takes a few seconds once before evaluation. Also, unlike the previous factorization-based methods (\eg IDC, GLaD) which still run pixel-level DD algorithms, LatentDD completely avoids computational overhead caused by repeatedly transforming components (\eg thumbnails, latent codes) into images and backpropagating gradient from images to the components, as the only method operating DD algorithms in latent space.
As for the space consumption, it is much more efficient to store latent codes into main memory than images when building datasets, and the low run-time GPU memory costs also allow LatentDD methods to run on higher resolution or greater data ratio. In conclusion, our LatentDD methods have settled the problems of high time \& space complexity (P1--P2) in \cref{sec:intro} as well.

\subsection{Ablation Studies}
\label{sec:ablation}

\begin{table*}[t]
    \centering
    \setlength{\tabcolsep}{1.9mm}
    \begin{tabular}{l|lccccc|lcc}
        \hline
        \multirow{2}{*}{Full set} & \multirow{2}{*}{Model} & \multicolumn{5}{c|}{Res. 256} & \multirow{2}{*}{Model} & \multicolumn{2}{c}{Res. 512} \\
        & & Bird & Fruit & Woof & Cat & Nette & & Bird & Fruit \\
        \hline
        Pixel ($\T$) & ConvNetD6 & 96.40 & 69.40 & 80.00 & 73.60 & 90.40 & ConvNetD7 & 92.20 & 68.80 \\
        Latent ($L^\T$, $f = 4$) & ConvNetD4 & 92.60 & 71.60 & 77.20 & 73.00 & 91.60 & ConvNetD5 & 95.20 & 75.60 \\
        Latent ($L^\T$, $f = 8$) & ConvNetD3 & 89.00 & 68.80 & 74.40 & 67.20 & 89.40 & ConvNetD4 & 92.80 & 72.40 \\
        \hline
    \end{tabular}
    \caption{Classification accuracy of the classifiers trained in pixel space and latent spaces respectively on the full sets without distillation.}
    \label{tab:full}
\end{table*}

\begin{table}[t]
\centering
\begin{tabular}{l|cc}
    \hline
     \multirow{2}{*}{Method} & \multicolumn{2}{c}{Bird 256} \\
     & IPC/LPC 1 & IPC/LPC 10 \\
     \hline
     DSA & 30.52 & 45.52 \\
     LatentDC & 29.68 & 45.56 \\
     \hline
     DM & 27.64 & 41.56 \\
     LatentDM & 26.64 & 45.00 \\
     \hline
     MTT & 35.80 & - \\
     LatentMTT & 33.60 & 46.88 \\
     \hline
\end{tabular}
\caption{Quantitative results of dataset distillation where both IPC and LPC are set to 1 or 10, and $f = 4$ for LatentDD methods.}
\label{tab:ipc1vslpc1}
\end{table}

\paragraph{Pixel Space vs. Latent Space}

We transfer the distillation processes from pixel space to latent space based on the fact that classification in latent space is approximately equivalent to that in pixel space. Along with the theoretical analysis in \cref{sec:latentspace}, we also empirically verify such equivalence by training classifiers directly on full datasets in pixel space $\T$ and latent spaces $L_\T$ without distillation, whose performance also represents the upper bound for DD methods. We may conclude from the results in \cref{tab:full} that
(1) the performance between the two spaces is generally comparable without significant disparity;
(2) when the resolution is lower (256), the latent space may be slightly inferior due to certain information loss;
(3) however when the resolution is high (512), the latent space outperforms the pixel space probably because there is too much redundant high-frequency information in pixel-level images and the information loss in latent codes decreases. Therefore, transferring classification tasks from pixel space to latent space is not only feasible, but may also bring some benefits by removing the useless information irrelevant for classification.

We also compare LatentDD methods under the same IPC/LPC rather than the same storage budget. As the results shown in \cref{tab:ipc1vslpc1}, LatentDD still achieves comparable performance with the DD algorithms in pixel space. These results also match \cref{fig:lossplot} where LatentDC can converge to a similar level of gradient matching loss to DC.

\paragraph{Downsampling Factor $f$ and Latents Per Class (LPC)}

In \cref{tab:f4vs8}, we additionally provide quantitative results of LatentDD methods with $f = 8$, where $\text{LPC} = 48 \times \text{IPC}$. As the number of latent codes has further increased, LatentDD methods reach even higher scores. However, as we have mentioned in \cref{sec:setting} that the autoencoder is originally pretrained on resolution 512, encoding and reconstructing images with overly low resolution and a large downsampling factor $f$ may induce too much loss of information that they may alter the distribution of the original images and finally counterweight the benefits brought by the quantity of latent codes (see examples and further analysis in Appendix). For instance, while reaching a mean accuracy of 44.38 when running LatentDC with $f = 4$ on \emph{Bird} at low resolution 64 and IPC 1, which is a comparable result at resolution 256 (46.72) or 512 (47.52) in \cref{tab:f4vs8}, we have observed a remarkable drop from 67.28/69.46 to 48.68 with $f = 8$. In practical applications, it is essential to emphasize the need for maintaining this balance according to the characteristics of specific datasets.

To verify that our LatentDD methods can indeed achieve performance gain by distilling datasets instead of merely delivering higher LPC than IPC within the same storage budget, latent codes initialized from randomly selected dataset images without distillation are also evaluated. See Appendix for results and analysis.

\begin{table}[t]
\centering
\begin{tabular}{lc|cc|cc}
     \hline
     \multirow{2}{*}{Method} & \multirow{2}{*}{$f$} & \multicolumn{2}{c|}{Res. 256 IPC 1} & \multicolumn{2}{c}{Res. 512 IPC 1} \\
     & & Bird & Fruit & Bird & Fruit  \\
     \hline
     \multirow{2}{*}{LatentDC} & 4 & 46.72 & 30.12 & 47.52 & 29.68 \\
     & 8 & 67.28 & 38.08 & 69.46 & 37.22 \\
     \hline
     \multirow{2}{*}{LatentDM} & 4 & 47.08 & 30.68 & 46.20 & 30.60 \\
     & 8 & 67.40 & 37.60 & 68.22 & 37.20\\
     \hline
     \multirow{2}{*}{LatentMTT} & 4 & 52.86 & 37.82 & 52.44 & 36.20 \\
     & 8 & 66.42 & 45.28 &  69.20 & 42.84 \\
     \hline
\end{tabular}
\caption{Quantitative results of dataset distillation where LatentDD methods are compared between $f = 4$ ($\text{LPC} = 12 \times \text{IPC}$) and $f = 8$ ($\text{LPC} = 48 \times \text{IPC}$).}
\label{tab:f4vs8}
\end{table}

\section{Conclusions}
\label{sec:conclusion}

In this work, we propose a new idea of transferring the dataset distillation (DD) processes from conventionally used pixel space to a more efficient latent space. Such transfer aims at three main problems commonly seen in the area of DD: \textbf{P1:} high time complexity; \textbf{P2:} high space complexity; and \textbf{P3:} low info-compactness, which have not been simultaneously settled in previous works. Practically, we propose LatentDD methods based on the three mainstream DD algorithms in pixel space by utilizing the latent codes encoded with a pretrained generic autoencoder as natural info-compact representations of pixel-level images. Experimental results have validated that LatentDD methods can not only largely reduce time and space consumption thus enable these methods on higher resolution or data ratio, but also significantly boost the performance in DD tasks by delivering quantitatively more info-compact latent codes than pixel-level images within the same storage budget.
{
    \small
    \bibliographystyle{ieeenat_fullname}
    \bibliography{latentdd}
}

\clearpage
\appendix
\section*{\Large Appendix}
\section{Implementation Detail}
\label{app:detail}

When running our LatentDD algorithms LatentDC, LatentDM and LatentMTT, we generally follow the default settings of the original DC, DM and MTT respectively. Besides, we fix some settings across the baselines and our methods to ensure that the comparisons are as fair as possible. In this section, we will introduce basic settings and implementation details of LatentDD, including 
\begin{itemize}
    \item hyperparameters in \cref{app:hyper},
    \item training and evaluation protocol in \cref{app:protocol},
    \item analysis of autoencoder \cref{app:ae},
    \item dataset selection and preprocessig in \cref{app:dataset},
    \item pesudocodes of the three LatentDD methods in \cref{app:code}.
\end{itemize}

\subsection{Hyperparameters}
\label{app:hyper}

\begin{table*}[t]
\centering
\begin{tabular}{c|c|c|c|ccc}
    \hline
     \multirow{5}{*}{Res.} & \multirow{5}{*}{IPC} & \multirow{5}{*}{$f$} & Hyperparameter &  LatentDC & LatentDM & LatentMTT \\
     \cline{4-7}
     & & & iteration $T$ & 1000 & 1000 & 5000 \\
     & & & max starting epoch $T^+$ & - & - & 5 \\
     & & & model learning rate $\eta_{\thetas}$ & 0.01 & - & 0.01 (initial)  \\
     & & & $\eta_{\thetas}$ learning rate $\lambda$ & - & - & $10^{-6}$ \\
     \hline
     \multirow{15}{*}{256} & \multirow{10}{*}{1} & \multirow{5}{*}{4} & outer loop / expert epoch $M$ & 10 & - & 1 \\
     & & & inner loop / student step $N$ & 50 & - & 40 \\
     & & & base learning rate $\eta_{\mathrm{base}}$ & 0.05 & 0.5 & 50 \\
     & & & batch size & 64 & 64 & 64 \\
     & & & ConvNet depth (train / eval) & 4 / 6 & 4 / 6 & 4 / 6 \\
     \cline{3-7}
     & &  \multirow{5}{*}{8} & outer loop / expert epoch $M$ & 10 & - & 1 \\
     & & & inner loop / student step $N$ & 50 & - & 60 \\
     & & & base learning rate $\eta_{\mathrm{base}}$ & 0.05 & 0.1 & 1 \\
     & & & batch size & 64 & 64 & 64 \\
     & & & ConvNet depth (train / eval) & 3 / 6 & 3 / 6 & 3 / 6 \\
     \cline{2-7}
     & \multirow{5}{*}{10} & \multirow{5}{*}{4} & outer loop / expert epoch $M$ & 10 & - & 1 \\
     & & & inner loop / student step $N$ & 50 & - & 80 \\
     & & & base learning rate $\eta_{\mathrm{base}}$ & 0.05 & 0.5 & 10 \\
     & & & batch size & 64 & 64 & 64 \\
     & & & ConvNet depth (train / eval) & 4 / 6 & 4 / 6 & 4 / 6 \\
     \hline
     \multirow{10}{*}{512} & \multirow{10}{*}{1} & \multirow{5}{*}{4} & outer loop / expert epoch $M$ & 10 & - & 1 \\
     & & & inner loop / student step $N$ & 50 & - & 40 \\
     & & & base learning rate $\eta_{\mathrm{base}}$ & 0.25 & 0.5 & 500 \\
     & & & batch size & 32 & 32 & 32 \\
     & & & ConvNet depth (train / eval) & 5 / 7 & 5 / 7 & 5 / 7 \\
     \cline{3-7}
     & & \multirow{5}{*}{8} & outer loop / expert epoch $M$ & 10 & - & 1 \\
     & & & inner loop / student step $N$ & 50 & - & 60 \\
     & & & base learning rate $\eta_{\mathrm{base}}$ & 0.05 & 0.1 & 20 \\
     & & & batch size & 16 & 16 & 16 \\
     & & & ConvNet depth (train / eval) & 4 / 7 & 4 / 7 & 4 / 7 \\
    \hline
\end{tabular}
\caption{Hyperparameter settings of the experiments on LatentDD algorithms. A hyphen - indicates that the hyperparameter is not applicable to this algorithm.}
\label{tab:hyperparam}
\end{table*}

The hyperparameters under different experimental settings are listed in \cref{tab:hyperparam}, except for other hyperparameters that have been introduced in our main paper or by previous works. Base learning rate $\eta_{\mathrm{base}}$ can be seen as the learning rate \emph{per latent code}, as the cross-entropy loss $\loss$ will be averaged among the latent codes. Hence the real learning rate $\eta_{L_\Ss}$ updating the latent codes is set to
\begin{equation}
    \eta_{L_\Ss} = \eta_{\mathrm{base}} \times \text{LPC}.
\end{equation}

Note that these settings are just used to produce the results reported in this work, they are not guaranteed to be the optimized settings to render the best performance.

\subsection{Training/Evaluation Protocol}
\label{app:protocol}

Following previous works, we adopt ConvNet \citep{convnet} as the network architecture primarily used in the experiments. By default, ConvNet of depth $d$ (denoted by ConvNetD$d$) consists of $d$ blocks followed by a single linear layer, where each block is a sequence of convolution with kernel size 3 and padding 1, instance normalization, ReLU activation and average pooling by factor 2. We set the depth $d$ of the ConvNet according to the resolution of the latent codes (during training stage) or pixel-level images (during evaluation stage), refer to \cref{tab:hyperparam} for details. The learning rate updating the network parameters during evaluation stage is set to 0.01.

As explained in the main paper, we do not apply Differentiable Siamese Augmentation (DSA) \cite{dsa} during the training processes because many augmenting transformations designed for pixel-level images do not fit for latent codes. However, we use DSA in the evaluation processes in pixel space. Following \citet{idc}, we additionally replace cutout with CutMix \citep{cutmix}, which is a calibration technique to alleviate the over-confident issue of the models trained on limited data.

\subsection{Analysis of Autoencoder}
\label{app:ae}

\begin{figure*}[t]
    \centering
    \includegraphics[width=.9\linewidth]{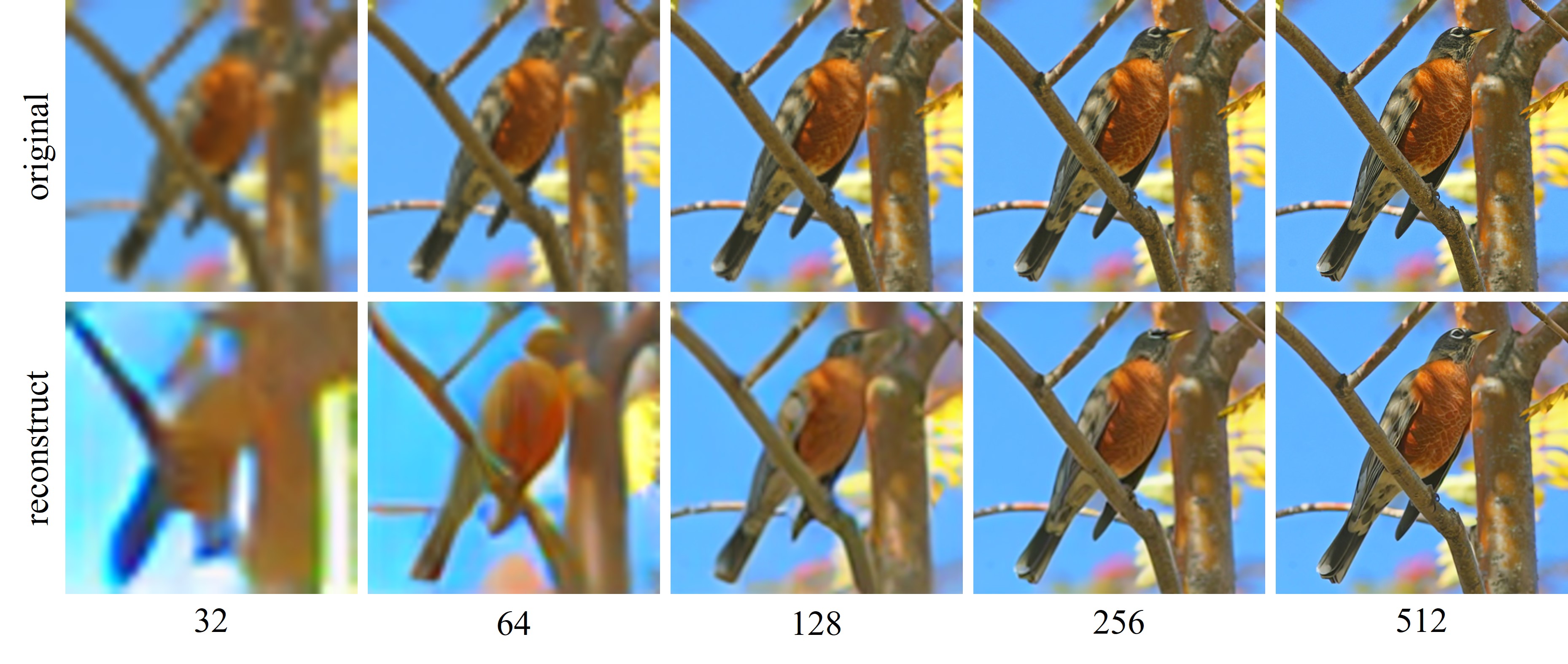}
    \caption{Comparison between the original images and the reconstructed images (encoded from the original images and decoded back by the autoencoder) at different resolutions from 32 to 512. Zoom in to see details of the high-resolution images.}
    \label{fig:recon}
\end{figure*}
\begin{table*}[t]
    \centering
    \setlength{\tabcolsep}{1.9mm}
    \begin{tabular}{l|ccccc|cc|cc}
        \hline
        \multirow{3}{*}{Method} & \multicolumn{7}{c|}{Res. 256} & \multicolumn{2}{c}{Res. 512} \\
        & \multicolumn{5}{c|}{IPC 1} & \multicolumn{2}{c|}{IPC 10} & \multicolumn{2}{c}{IPC 1} \\
        & Bird & Fruit & Woof & Cat & Nette & Bird & Fruit & Bird & Fruit \\
        \hline
        Initial & 32.36 & 23.24 & 20.20 & 29.04 & 43.72 & 75.28 & 45.80 & 32.56 & 22.96 \\
        LatentDC & 46.72 & 30.12 & 28.96 & 38.08 & 55.92 & 80.44 & 51.60 & 47.52 & 29.68 \\
        LatentDM & 47.08 & 30.68 & 28.00 & 36.28 & 56.08 & 77.20 & 47.76 & 46.20 & 30.60 \\
        LatentMTT & 52.86 & 37.82 & 39.84 & 41.42 & 62.86 & 78.44 & 52.46 & 52.44 & 36.20 \\
        \hline
    \end{tabular}
    \caption{Comparison between the latent codes before distillation (Initial), and after running LatentDD algorithms on ImageNet subsets, following the setting of $f = 4$ and $\text{LPC} = 12 \times \text{IPC}$. The mean classification accuracy among five times of evaluations is reported.}
    \label{tab:init}
\end{table*}
\begin{table}[t]
    \centering
    \setlength{\tabcolsep}{1.5mm}
    \begin{tabular}{l|cccc}
        \hline
        \multirow{2}{*}{Method} & \multicolumn{4}{c}{Bird 256 IPC 1} \\
        & ConvNet & ResNet18 & VGG11 & AlexNet \\
        \hline
        DSA & 30.52 & 14.84 & 20.68 & 24.24 \\
        IDC & 36.28 & 40.24 & 30.20 & 32.24 \\
        GLaD DC & 30.32 & 23.20 & 22.08 & 19.44 \\
        \textbf{LatentDC} & \textbf{46.72} & \textbf{56.00} & \textbf{49.32} & \textbf{37.56} \\
        \hline
        DM & 27.64 & 13.44 & 17.40 & 23.64 \\
        GLaD DM & 28.84 & 26.24 & 18.24 & 19.08 \\
        \textbf{LatentDM} & \textbf{47.08} & \textbf{56.00} & \textbf{47.56} & \textbf{37.12} \\
        \hline
        MTT & 35.80 & 18.72 & 22.36 & 20.84 \\
        GLaD MTT & 32.20 & 39.56 & 23.28 & 19.80 \\
        FTD & 35.96 & 19.28 & 21.96 & 23.92 \\
        \textbf{LatentMTT} & \textbf{52.86} & \textbf{57.76} & \textbf{52.96} & \textbf{39.64} \\
        \hline
    \end{tabular}
    \caption{Quantitative results of dataset distillation experiments on ImageNet subset \textit{Bird}, evaluated on cross-architecture networks (ResNet18, VGG11 and AlexNet). LatentDD algorithms follow the setting of $f = 4$, where $\text{LPC} = 12 \times \text{IPC}$. The mean classification accuracy among five times of evaluations is reported.}
    \label{tab:crossarch}
\end{table}

In \cref{fig:recon} we show some examples of reconstructing original images at different resolutions with the autoencoder used in LatentDD, with $f = 8$ and without pre-upsampling and post-downsampling. Although the autoencoder can be applied to a variety of resolutions, we observe that the lower the resolution is, the more details will be lost in the reconstruction because the resolution of the latent codes becomes too small to encode spatial information. 
While the reconstructed image of resolution 512 only loses the finest details that are hardly perceptible, the image of resolution 32 loses too much information to even tell that it depicts a bird.
Therefore, the experiments in our paper are mainly rolled out with $f = 4$ where we adopt the (1) upsampling; (2) encoding; (3) LatentDD; (4) decoding; and (5) downsampling procedure for a generalized applicability on various resolutions, though in high-resolution scenarios $f = 8$ may produce better results due to the greater quantity of latent codes. In extreme cases where the dataset resolution is too low, we suggest trying $f = 2$ with pre-upsampling and post-downsampling by factor of 4 to avoid excessive loss of information, or replace with another pretrained generic autoencoder which suits low-resolution images. For other data modality (\eg, text, audio, video) where autoencoders may not be available, the idea of LatentDD is still applicable if there is a way to compress the data into smaller size without altering the dataset distribution and reconstruct them back to the original form, depending on the specific modality.

\subsection{Dataset Selection and Preprocessing}
\label{app:dataset}

We list the categories (and their indices) that are included in the ImageNet \cite{imagenet} subsets below.
\begin{itemize}
    \item \textbf{Bird:} peacock (84), flamingo (130), macaw (88), pelican (144), king penguin (145), bald eagle (22), toucan (96), ostrich (9), black swan (100), sulphur-crested cockatoo (89).
    \item \textbf{Fruit:} pineapple (953), banana (954), strawberry (949), orange (950), lemon (951), pomegranate (957), fig (952), bell pepper (945), cucumber (943), granny smith (948).
    \item \textbf{Woof:} australian terrier (193), border terrier (182), samoyed (258), beagle (162), shih-tzu (155), english foxhound (167), rhodesian ridgeback (159), dingo (273), golden retriever (207), old english sheepdog (229).
    \item \textbf{Cat:} tabby (281), tiger cat (282), persian cat (283), siamese cat (284), egyptian cat (285), lion (291), tiger (292), jaguar (290), snow leopard (289), lynx (287).
    \item \textbf{Nette:} tench (0), english springer (217), cassette player (482), chain saw (491), church (497), french horn (566), garbage truck (569), gas pump (571), golf ball (574), parachute (701).
\end{itemize}

For preprocessing the images of ImageNet, we follow the previous works \cite{mtt,glad} to apply a sequence of (1) normalization; (2) resizing to target resolution; and (3) center cropping.

\subsection{Pseudocodes of LatentDD Algorithms}
\label{app:code}

The pseudocodes of the three LatentDD methods, LatentDC, LatentDM and LatentMTT are respectively shown in Algorithm \ref{algo:latentdc}, \ref{algo:latentdm} and \ref{algo:latentmtt}.

\section{Additional Results}
\label{app:result}

In this section, we illustrate some additional experimental results that have not been shown in the main paper due to page limitation. These results include
\begin{itemize}
    \item the evaluation on the initial latent codes encoded from randomly selected dataset images before distillation in \cref{app:init},
    \item the performance of LatentDD methods when evaluated on cross-architecture networks in \cref{app:crossarch},
    \item some image samples decoded from the distilled latent codes in \cref{app:qual}.
\end{itemize}

\subsection{Latent Codes Before vs. After Distillation}
\label{app:init}

In \cref{tab:init} (row \textit{Initial}) we list the classification accuracy of the latent codes initialized from random dataset images before distillation. Compared with the results of LatentDD algorithms, we may conclude that our LatentDD methods do not achieve the best performance only because they deliver more latent codes than pixel-level images within the same storage budget. Instead they still improve the latent codes during the distillation processes, albeit the performance gain is relatively small when $\text{IPC} = 10$ probably because the potential for further improvement is limited when we already have $120$ latent codes per class.

\subsection{Cross-architecture Performance}
\label{app:crossarch}

One of the goals of dataset distillation is that we expect the distilled datasets can achieve good performance on any network architecture, not limited to the one adopted during distillation, so that these distilled datasets can be utilized on downstream tasks such as neural architecture search \cite{nas}. In \cref{tab:crossarch}, we demonstrate the uniformly good performance of LatentDD algorithms on some cross-architecture networks. As all the baselines are trained on the same architecture used for evaluation (\eg ConvNetD6 for resolution 256), they are more or less overfitted to the specific architecture as a trend of performance drop can be observed when evaluated on other architectures. On the contrary, our LatentDD methods are relatively robust to architecture changes. It is also worth mentioning that the results of LatentDD methods on ConvNet in \cref{tab:crossarch} (and also the tables in the main paper which include performance of LatentDD) are already cross-architecture to some extent, since the depths of the ConvNets used for training and evaluation are different.

\subsection{Qualitative Results}
\label{app:qual}

\begin{figure*}[p]
    \centering
    \includegraphics[width=\linewidth]{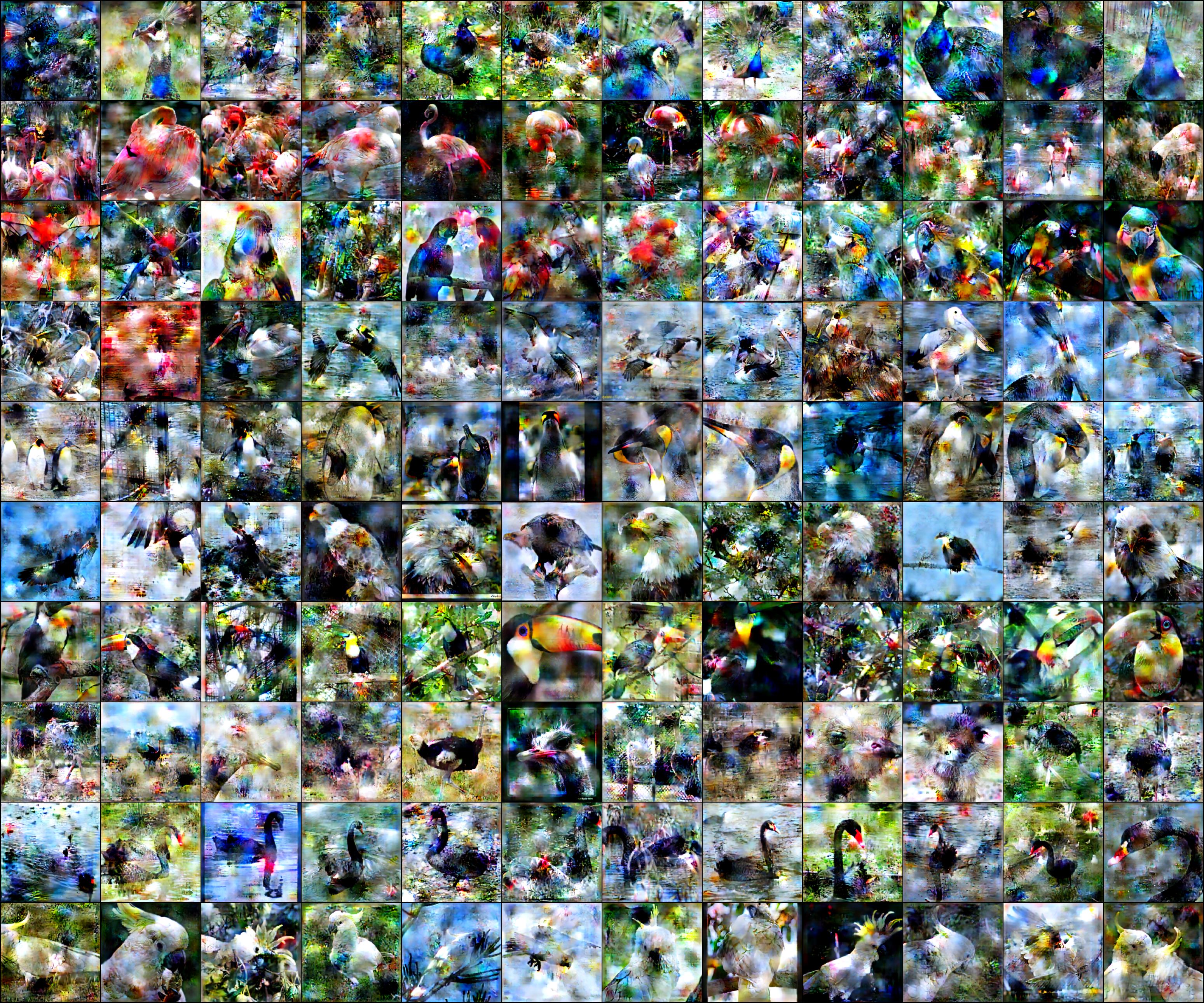}
    \caption{The images decoded from the distilled latent codes by \textbf{LatentDC} with $f = 4$ on \emph{Bird}, resolution 256, IPC 1 (LPC 12). Each row respectively presents the category of peacock, flamingo, macaw, pelican, king penguin, bald eagle, toucan, ostrich, black swan, sulphur-crested cockatoo.}
    \label{fig:quallatentdc}
\end{figure*}

\begin{figure*}[p]
    \centering
    \includegraphics[width=\linewidth]{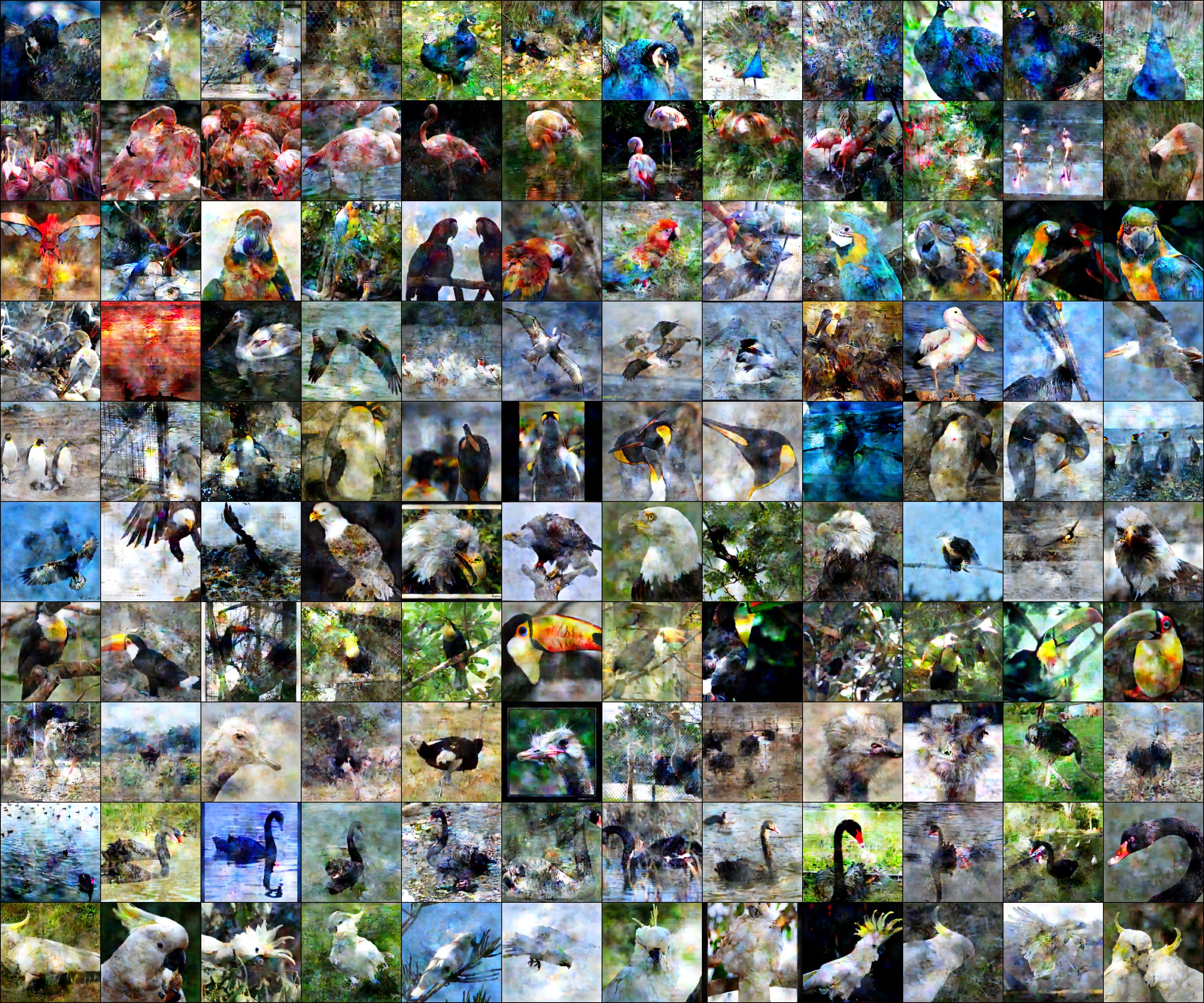}
    \caption{The images decoded from the distilled latent codes by \textbf{LatentDM} with $f = 4$ on \emph{Bird}, resolution 256, IPC 1 (LPC 12). Each row respectively presents the category of peacock, flamingo, macaw, pelican, king penguin, bald eagle, toucan, ostrich, black swan, sulphur-crested cockatoo.}
    \label{fig:quallatentdm}
\end{figure*}

\begin{figure*}[p]
    \centering
    \includegraphics[width=\linewidth]{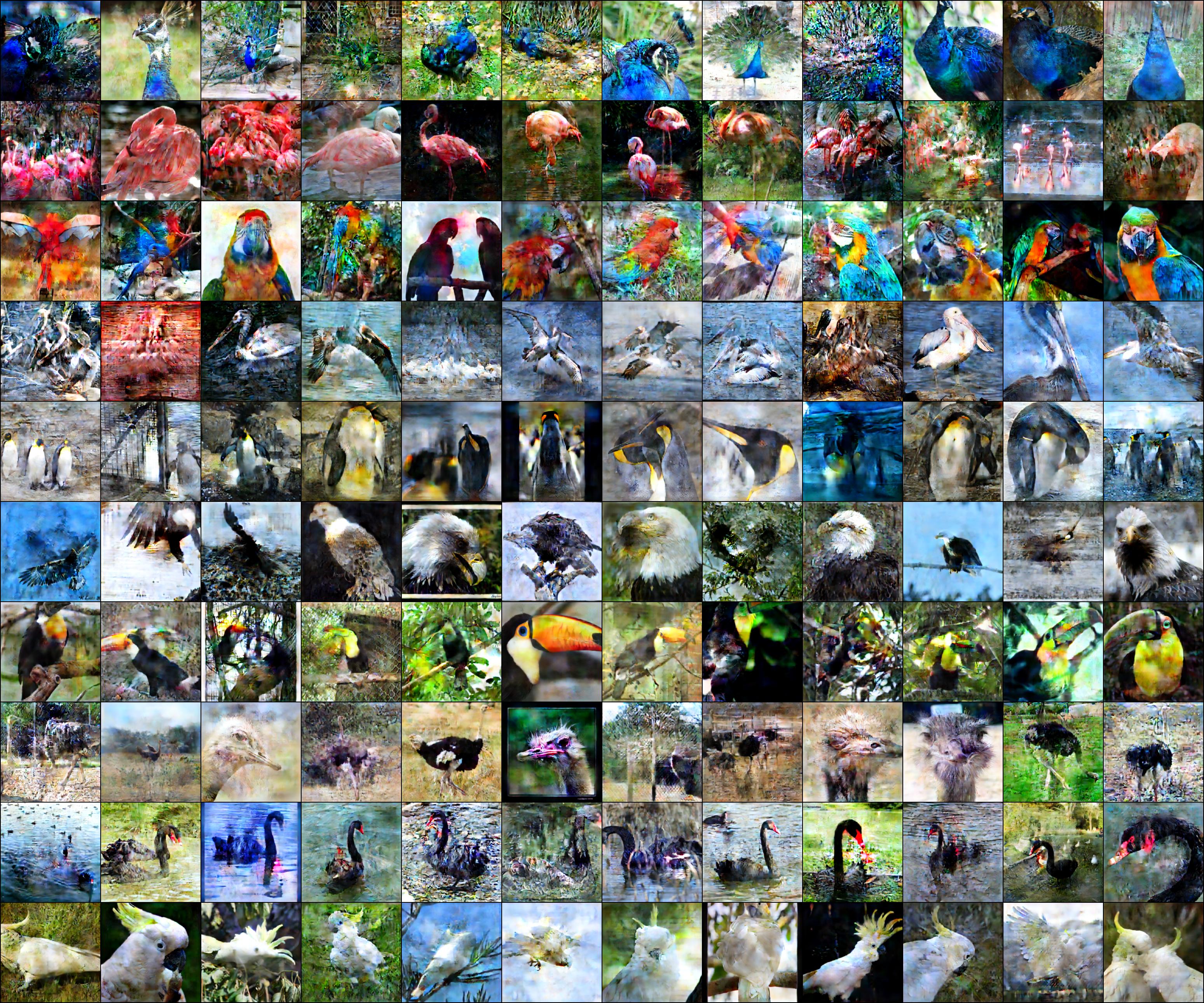}
    \caption{The images decoded from the distilled latent codes by \textbf{LatentMTT} with $f = 4$ on \emph{Bird}, resolution 256, IPC 1 (LPC 12). Each row respectively presents the category of peacock, flamingo, macaw, pelican, king penguin, bald eagle, toucan, ostrich, black swan, sulphur-crested cockatoo.}
    \label{fig:quallatentmtt}
\end{figure*}

For a more comprehensive illustration of the experimental results, we depict some image samples decoded from the distilled latent codes on the ImageNet subset \emph{Bird} (with categories listed in \cref{app:dataset}) by LatentDC, LatentDM and LatentMTT in \cref{fig:quallatentdc,fig:quallatentdm,fig:quallatentmtt} respectively.

\begin{algorithm*}[p]
\small
\KwIn{Real dataset in latent space $L_\T = \{(l_i, y_i)\}_{i=1}^{|\T|}$ where $l_i = \enc(x_i)$; iteration $T$; outer loop $M$; inner loop $N$; number of classes $C$; classification loss $\ce(\cdot, \cdot)$; gradient matching loss $D(\cdot, \cdot)$; learning rate of synthetic dataset $\eta_{L_\Ss}$ and model parameters $\eta_{\thetas}$. }

Initialize synthetic dataset in latent space $L_\Ss = \{(\ls_i, \ys_i)\}_{i=1}^{|\Ss|}$ with randomly sampled $(l_i, y_i) \in L_\T$.

\For{$\mathrm{iter} = 0, \dots, T - 1 \mathcal{}$} 	
{
    Initialize model $\phi_{\thetas}$ with random parameters $\thetas$.
    
    \For{$m = 0, \dots, M - 1$}
    {
        $\loss_{\mathrm{match}} = 0$
    
        \For{$c = 0, \dots, C - 1$}
        {
            Randomly sample a real batch $B_{L_\T} \subseteq L_{\T_c}$ and a synthetic batch $B_{L_\Ss} \subseteq L_{\Ss_c}$ of class $c$. 
            
            $\loss_c^{L_\T}(\thetas) = \frac{1}{|B_{L_\T}|} \sum_{(l,y) \in B_{L_\T}}\ce(\phi_{\thetas}(l), y)$
            
            $\loss_c^{L_\Ss}(\thetas) = \frac{1}{|B_{L_\Ss}|} \sum_{(\ls,\ys) \in B_{L_\Ss}}\ce(\phi_{\thetas}(\ls), \ys)$
            
            $\loss_{\mathrm{match}} \gets \loss_{\mathrm{match}} + D(\nabla_{\thetas} \loss_c^{L_\Ss}(\thetas), \nabla_{\thetas} \loss_c^{L_\T}(\thetas))$
        }
        
        $L_\Ss \gets L_\Ss - \eta_{L_\Ss} \nabla_{L_\Ss} \loss_{\mathrm{match}}$ \tcp{update synthetic latent codes with gradient matching}
        
        \For{$n = 0, \dots, N - 1$}
        {
            $\thetas \gets \thetas - \eta_{\thetas} \nabla_{\thetas} \loss^{L_\T}(\thetas)$ \tcp{train the model on real latent codes}
        }
    }
}
\KwOut{$L_\Ss$}
\caption{LatentDC}
\label{algo:latentdc}
\end{algorithm*}
\begin{algorithm*}[p]
\small
\KwIn{Real dataset in latent space $L_\T = \{(l_i, y_i)\}_{i=1}^{|\T|}$ where $l_i = \enc(x_i)$; iteration $T$; learning rate of synthetic dataset $\eta_{L_\Ss}$. }

Initialize synthetic dataset in latent space $L_\Ss = \{(\ls_i, \ys_i)\}_{i=1}^{|\Ss|}$ with randomly sampled $(l_i, y_i) \in L_\T$.

\For{$\mathrm{iter} = 0, \dots, T - 1 \mathcal{}$} 	
{
    Initialize model $\phi_{\thetas}$ with random parameters $\thetas$.
    
    $\loss_{\mathrm{match}} = 0$
    
    \For{$c = 0, \dots, C - 1$}
    {
        Randomly sample a real batch $B_{L_\T} \subseteq L_{\T_c}$ and a synthetic batch $B_{L_\Ss} \subseteq L_{\Ss_c}$ of class $c$. 
        
        $\loss_{\mathrm{match}} \gets \loss_{\mathrm{match}} + \|\frac{1}{|B_{L_\T}|} \sum_{(l, y) \in B_{L_\T}} \phi_{\thetas}(l) - \frac{1}{|B_{L_\Ss}|} \sum_{(\ls, \ys) \in B_{L_\Ss}} \phi_{\thetas}(\ls) \|^2$
    }
    
    $L_\Ss \gets L_\Ss - \eta_{L_\Ss} \nabla_{L_\Ss} \loss_{\mathrm{match}}$ \tcp{update synthetic latent codes with distribution matching}
}
\KwOut{$L_\Ss$}
\caption{LatentDM}
\label{algo:latentdm}
\end{algorithm*}
\begin{algorithm*}[p]
\small
\KwIn{Real dataset in latent space $L_\T = \{(l_i, y_i)\}_{i=1}^{|\T|}$ where $l_i = \enc(x_i)$; iteration $T$; maximum starting epoch $T^+$; expert epoch $M$; student step $N$; number of classes $C$; classification loss $\ce(\cdot, \cdot)$; learning rate of synthetic dataset $\eta_{L_\Ss}$; initial learning rate of model parameters $\eta_{\thetas}^0$; learning rate of updating the the learning rate of model parameters $\lambda$. }

Initialize synthetic dataset in latent space $L_\Ss = \{(\ls_i, \ys_i)\}_{i=1}^{|\Ss|}$ with randomly sampled $(l_i, y_i) \in L_\T$.

Initialize learning rate of model parameters $\eta_{\thetas} = \eta_{\thetas}^0$.

\For{$\mathrm{iter} = 0, \dots, T - 1 \mathcal{}$} 	
{
    Randomly sample a starting epoch $t \sim \mathcal{U}(\{0, \dots, T^+\})$.
    
    Randomly choose an expert trajectory from the buffer: $\thetas_{t}^{L_\T}, \dots, \thetas_{t + M}^{L_\T}$.
    
    Initialize student network $\phi$ with parameters $\thetas = \thetas_{t}^{L_\T}$.
    
    \For{$n = 0, \dots, N - 1$}
    {
        Randomly sample a synthetic batch $B_{L_\Ss} \subseteq L_{\Ss}$.
        
         $\loss^{L_\Ss}(\thetas) = \frac{1}{|B_{L_\Ss}|} \sum_{(\ls,\ys) \in B_{L_\Ss}}\ce(\phi_{\thetas}(\ls), \ys)$
         
         $\thetas \gets \thetas - \eta_{\thetas} \nabla_{\thetas} \loss^{L_\Ss}(\thetas)$ \tcp{student step}
    }
    
    $\loss_{\mathrm{match}} = \|\thetas - \thetas_{t + M}^{L_\T}\|_2^2 / \|\thetas_{t}^{L_\T} - \thetas_{t + M}^{L_\T}\|_2^2$
    
    $L_\Ss \gets L_\Ss - \eta_{L_\Ss} \nabla_{L_\Ss} \loss_{\mathrm{match}}$ \tcp{update synthetic latent codes with trajectory matching}
 
    $\eta_{\thetas} \gets \eta_{\thetas} - \lambda \nabla_{\eta_{\thetas}} \loss_{\mathrm{match}}$
}
\KwOut{$L_\Ss$}
\caption{LatentMTT}
\label{algo:latentmtt}
\end{algorithm*}

\end{document}